\documentclass[10pt,twocolumn,a4paper]{esaAI}
\usepackage[table,xcdraw]{xcolor}
\usepackage{multirow}
\usepackage{booktabs}
\usepackage{float}

\title{AstroSpy: On detecting Fake Images in Astronomy via Joint Image-Spectral Representations}

\def\authorEmail{mohammed.alam@mbzuai.ac.ae}

\author[1]{Mohammed Talha Alam\thanks{Corresponding author: \authorEmail}\thanks{Equal Contribution}}
\author[1]{Raza Imam$^{\dag}$}
\author[1]{Mohsen Guizani}
\author[1]{Fakhri Karray}
\affil[1]{Mohamed bin Zayed University of Artificial Intelligence, Abu Dhabi, UAE}

\begin{document}

\makeCustomtitle

\begin{abstract}
The prevalence of AI-generated imagery has raised concerns about the authenticity of astronomical images, especially with advanced text-to-image models like Stable Diffusion producing highly realistic synthetic samples. Existing detection methods, primarily based on convolutional neural networks (CNNs) or spectral analysis, have limitations when used independently. We present \texttt{AstroSpy}, a hybrid model that integrates both spectral and image features to distinguish real from synthetic astronomical images. Trained on a unique dataset of real NASA images and AI-generated fakes (approximately 18k samples), \texttt{AstroSpy} utilizes a dual-pathway architecture to fuse spatial and spectral information. This approach enables \texttt{AstroSpy} to achieve superior performance in identifying authentic astronomical images. Extensive evaluations demonstrate \texttt{AstroSpy}'s effectiveness and robustness, significantly outperforming baseline models in both \textit{in-domain} and \textit{cross-domain} tasks, highlighting its potential to combat misinformation in astronomy.
\end{abstract}
\section{Introduction}
The authenticity of visual data is paramount in scientific fields, particularly in astronomy, where images are central to research, discovery, and public engagement. However, the advent of sophisticated AI models capable of generating highly realistic images has led to a surge in fake astronomical visuals. These fake images can deceive the public, mislead researchers, and potentially disrupt scientific communication. The challenge lies in developing robust methods to discern real images from AI-generated fakes, ensuring the integrity of visual data \cite{rafique2023deep, bird2024cifake, heidari2024deepfake, yan2023ucf}. Given the increasing realism of these fake images, the need for robust detection methods is more urgent than ever \cite{shamshad2023evading}.
\vspace{-1.0em}
\paragraph{Motivation}
Misleading images can undermine public trust in scientific findings and hinder scientific progress \cite{nguyen2022deep, imam2023optimizing, hackstein2023evaluation}. This can affect the allocation of resources and funding. For instance, if AI-generated images depicting spectacular but fictitious celestial phenomena capture the public's imagination, funding bodies might divert resources away from legitimate scientific research to capitalize on the public interest. This shift can undermine genuine scientific inquiry and the development of accurate astronomical knowledge. Additionally, generated images, while potentially valuable for outreach, must be clearly distinguished from actual data to maintain scientific integrity \cite{kim2024generative}. The artistic interpretation of astronomical data, such as infrared imagery, already involves some level of abstraction and creativity, but it is grounded in real observations and scientific analysis, which differs fundamentally from wholly fabricated images \cite{smith2023astronomia}. 


\vspace{-1.2em}
\paragraph{Related Works}
Detecting fake images has been extensively studied in digital forensics and deepfake detection, with CNNs being pivotal due to their ability to learn complex image features \cite{razzak2018deep, rana2022deepfake, masood2023deepfakes}. However, these models often struggle with sophisticated forgeries. Spectral analysis, which examines frequency domain features, has shown promise in detecting anomalies \cite{farid2009image, cozzolino2015splicebuster, cheng2024deep}. Durall et al. \cite{durall2020watch} emphasized the importance of frequency domain analysis in detecting GAN-generated images, noting that many GANs fail to reproduce the spectral distributions of real images accurately. Similarly, \cite{frank2020leveraging} leveraged frequency domain information to detect manipulated images. Our work bridges these approaches by combining spatial and spectral features to address the unique challenges of detecting fake astronomical imagery.
\vspace{-1.2em}
\begin{figure*}[!t]
    \centering
    \includegraphics[width=0.90\textwidth]{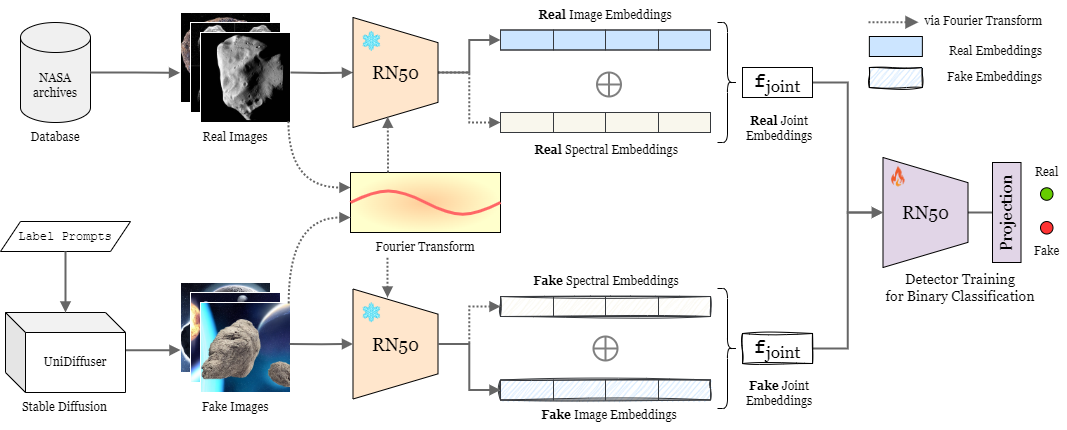}
    \caption{\texttt{AstroSpy} architecture: Real images from database archive, and fake images generated by Stable Diffusion, are transformed into spectral embeddings using Fourier Transform. Both image and spectral features are extracted using ResNet50, concatenated into joint embeddings, and used to train a binary classifier for detecting real and fake images.}
    \label{fig:enter-label}
\end{figure*}
\paragraph{Contribution}
Recognizing the limitations of these approaches when used in isolation, (1) we introduce \texttt{AstroSpy}, a novel model that integrates both spectral and image features to enhance the detection of fake astronomical images. (2) We utilize a robust dataset composed of real astronomical images from NASA and AI-generated fakes sourced from the \textit{`Raw\_Aug\_HR'} and \textit{`T2I\_Aug\_HR'} datasets within the FLARE \cite{alam2024flare} framework. This diverse dataset is crucial for effectively training and evaluating our detection models. (3) Through extensive experiments, including the creation of \textit{out-of-domain} datasets using Unidiffuser \cite{bao2023one}, Roentgen \cite{chambon2022roentgen}, and StyleGAN2 \cite{karras2020analyzing} for natural images, medical images, and faces, we demonstrate that \texttt{AstroSpy} significantly outperforms existing methods in both \textit{in-domain} and \textit{cross-domain} scenarios.

\section{Method: \texttt{AstroSpy}}
\texttt{AstroSpy}'s workflow leverages a novel combination of image and spectral features to detect fake astronomical images.
\vspace{-0.40em}
\subsection{Dataset Preparation}
We extracted real astronomical images from NASA's archives using keywords such as planet, star, nebula, galaxy, constellation, black hole, and asteroid, followed by different augmentations, making a comprehensive set of about $\sim$9k real samples. For generating fake samples, we utilized a multi-modal stable diffusion method \cite{bao2023transformer}. By prompting the class labels, we generate $\sim$9k synthetic images that closely mimic the context and visual content of the real samples.
\vspace{-0.40em}
\subsection{Joint Embeddings}
\textbf{Image Features}: We utilize a base ResNet50 encoder for extracting image features. ResNet50 is renowned for its deep architecture and ability to capture intricate visual patterns \cite{he2016deep}. Let $I$ represent the input image, and $F_{\text{ResNet50}}(I)$ denote the feature vector obtained from the ResNet50 encoder:
\begin{equation}
    \mathbf{f}_{\text{image}} = F_{\text{ResNet50}}(I)
\end{equation}
\textbf{Spectral Features}: For spectral analysis, grayscale images undergo a Fourier transform \cite{duoandikoetxea2024fourier} to convert them into the frequency domain. Let $G(I)$ represent the grayscale image derived from $I$, and $\mathcal{F}(G(I))$ denote the Fourier transform of the grayscale image. The magnitude spectrum of the Fourier transform is computed, and these spectral features are processed using the ResNet50 encoder. Let $F_{\text{ResNet50}}(\mathcal{M})$ denote the feature vector obtained from the magnitude spectrum $\mathcal{M}$ of the Fourier transform:
\begin{equation}
    \mathbf{f}_{\text{spectral}} = F_{\text{ResNet50}}(\mathcal{M}), \quad \text{where} \quad \mathcal{M} = |\mathcal{F}(G(I))|
\end{equation}
The final joint embedding $\mathbf{f}_{\text{joint}}$ is obtained by concatenating the image and spectral feature vectors:
\begin{equation}
    \mathbf{f}_{\text{joint}} = [\mathbf{f}_{\text{image}} \odot \mathbf{f}_{\text{spectral}}]
    \label{eq:joint_embeds}
\end{equation}
where $\odot$ indicates a concatenation function. For \texttt{AstroSpy}, we concatenated $\mathbf{f}_{image}$ and $\mathbf{f}_{spectral}$ using addition operation. This joint embedding $\mathbf{f}_{joint}$ is then fed into a trainable model (like ResNet50 or DenseNet121) followed by fully connected layer for binary classification:
\begin{equation}
    \hat{y} = \sigma(\mathbf{W} \mathbf{f}_{\text{joint}} + \mathbf{b})
\end{equation}
where $\sigma$ represents the sigmoid activation function, $\mathbf{W}$ is the weight matrix, $\mathbf{b}$ is the bias vector, and $\hat{y}$ is the predicted probability of the input image being real.


\subsection{Training Procedure}

\texttt{AstroSpy} utilizes the binary cross-entropy loss used to measure the discrepancy between the predicted probabilities $\hat{y}$ and the actual labels $y$ as:
    \vspace{-0.10em}
    \begin{equation}
        \mathcal{L} = -[y \log(\hat{y}) + (1 - y) \log(1 - \hat{y})]
    \end{equation}
    \vspace{-0.10em}
    The gradients of the loss function with respect to the model parameters are computed and used to update the parameters iteratively using the Adam optimizer:
    \begin{equation}
        \theta \leftarrow \theta - \eta \nabla_{\theta} \mathcal{L}
    \end{equation}
    \vspace{-0.25em}
    where $\theta$ represents the model parameters, $\eta$ is the learning rate, and $\nabla_{\theta} \mathcal{L}$ is the gradient of the loss function with respect to the parameters.
Through this training, \texttt{AstroSpy} learns to accurately distinguish between real and fake astronomical images.



{\renewcommand{\arraystretch}{1.0}
\begin{table*}[htp]
\caption{Performance when trained on different embedding combinations. \textbf{Joint} indicates concatenated embeddings $\mathbf{f}_{joint}$.}
\vspace{-0.75em}
\centering
\resizebox{\textwidth}{!}{%
\begin{tabular}{l|l|lll|l|l}
\hline
\multicolumn{1}{c|}{\cellcolor[HTML]{EFEFEF}} & \multicolumn{1}{c|}{\cellcolor[HTML]{EFEFEF}\textbf{In-Domain}} & \multicolumn{4}{c}{\cellcolor[HTML]{EFEFEF}\textbf{Out-Of-Domain (OOD)}} & \multicolumn{1}{|c}{\cellcolor[HTML]{EFEFEF}} \\ 
\rowcolor[HTML]{EFEFEF} 
\multicolumn{1}{c|}{\multirow{-2}{*}{\cellcolor[HTML]{EFEFEF}\textbf{Configuration}}} & \multicolumn{1}{c|}{\cellcolor[HTML]{EFEFEF}\textbf{Real$\texttt{+}$Fake}} & \multicolumn{1}{c}{\cellcolor[HTML]{EFEFEF}\textbf{ImageNet}} & \multicolumn{1}{c}{\cellcolor[HTML]{EFEFEF}\textbf{MIMIC}} & \multicolumn{1}{c}{\cellcolor[HTML]{EFEFEF}\textbf{Fairface }} & \multicolumn{1}{c}{\cellcolor[HTML]{EFEFEF}\textbf{OOD Average}} & \multicolumn{1}{|c}{\multirow{-2}{*}{\cellcolor[HTML]{EFEFEF}\textbf{Average}}} \\ \hline

Fourier embeds (bs.) & 0.69 & 0.59 & 0.55 & 0.51 & 0.55 & 0.58 \\
Img embeds & 0.89 & 0.64 & 0.50 & 0.70 & 0.61 & 0.68 \\
Img embeds + Augs. & 0.94 & 0.71 & 0.55 & 0.56 & 0.60 & 0.69 \\
\rowcolor[HTML]{E4EDFA} 
\textbf{Joint (\texttt{AstroSpy})} & \textbf{0.98}$~\uparrow$ & \textbf{0.83}$~\uparrow$ & \textbf{0.59}$~\uparrow$ & \textbf{0.93}$~\uparrow$ & \textbf{0.78}$~\uparrow$ & \textbf{0.83}$~\uparrow$\\

\hline
\end{tabular}}
\label{tab:results}
\end{table*}}

{\renewcommand{\arraystretch}{1.0}
\begin{table*}[htp]
\caption{Classification performance of \texttt{AstroSpy} when trained across different backbones.}
\vspace{-0.75em}
\centering
\resizebox{\textwidth}{!}{%
\begin{tabular}{l|l|lll|l|l}
\hline
\rowcolor[HTML]{EFEFEF} 
\multicolumn{1}{c|}{\cellcolor[HTML]{EFEFEF}} & \multicolumn{1}{c|}{\cellcolor[HTML]{EFEFEF}\textbf{In-Domain}} & \multicolumn{4}{c}{\cellcolor[HTML]{EFEFEF}\textbf{Out-Of-Domain (OOD)}} & \multicolumn{1}{|c}{\cellcolor[HTML]{EFEFEF}} \\
\rowcolor[HTML]{EFEFEF} 
\multicolumn{1}{c|}{\multirow{-2}{*}{\cellcolor[HTML]{EFEFEF}\textbf{Trained Backbone}}} & \multicolumn{1}{c|}{\cellcolor[HTML]{EFEFEF}\textbf{Real$\texttt{+}$Fake}} & \multicolumn{1}{c}{\cellcolor[HTML]{EFEFEF}\textbf{ImageNet}} & \multicolumn{1}{c}{\cellcolor[HTML]{EFEFEF}\textbf{MIMIC}} & \multicolumn{1}{c}{\cellcolor[HTML]{EFEFEF}\textbf{Fairface }} & \multicolumn{1}{c}{\cellcolor[HTML]{EFEFEF}\textbf{OOD Average}} & \multicolumn{1}{|c}{\multirow{-2}{*}{\cellcolor[HTML]{EFEFEF}\textbf{Average}}} \\ \hline

CNN & 0.82 & 0.53 & 0.48 & 0.51 & 0.50 & 0.58 \\
DenseNet121 & 0.96 & 0.71 & 0.55 & 0.82 & 0.69 & 0.76 \\
\rowcolor[HTML]{E4EDFA} 
\textbf{ResNet50} & \textbf{0.98}$~\uparrow$ & \textbf{0.83}$~\uparrow$ & \textbf{0.59}$~\uparrow$ & \textbf{0.93}$~\uparrow$ & \textbf{0.78}$~\uparrow$ & \textbf{0.83}$~\uparrow$ \\

\hline
\end{tabular}}
\label{tab:backbone}
\end{table*}}


\section{Results and Discussion}
\subsection{Experiments}

\paragraph{Datasets} We use a curated dataset from \cite{alam2024flare}, containing approximately 9k real samples and 9k synthetic samples across eight classes: Planet, Asteroid, Nebula, Comet, Star, Black Hole, Galaxy, and Constellation. The real samples are sourced from NASA's archives, while synthetic samples are generated using Stable Diffusion \cite{rombach2022high}. The dataset is split into training, validation, and test sets in an 80-10-10 ratio. For \textit{out-of-domain} evaluation, we included real samples from ImageNet (natural images) \cite{deng2009imagenet}, MIMIC (medical images) \cite{johnson2019mimic}, and FairFace (face images) \cite{karkkainen2021fairface}, with corresponding synthetic samples generated by UniDiffuser, Roentgen, and StyleGAN2, respectively.
\vspace{-1.0em}
\paragraph{Implementation Details}
\texttt{AstroSpy} is implemented using PyTorch \cite{paszke2019pytorch} with two parallel ResNet50 models, one for image features and one for spectral features. Images are resized to 224×224 pixels for both pathways. The final fully connected layer of each ResNet50 is replaced with an identity layer to use the output as feature vectors. The model is trained with a batch size of 32, using the Adam optimizer at a learning rate of 2e-5, and cross-entropy loss for binary classification. Data augmentation techniques such as random rotations, flips, color jittering, and Gaussian blurring are applied to enhance robustness. Early stopping with a patience of 5 epochs is employed to prevent overfitting. Training is conducted for 25 epochs on a single NVIDIA A100 40GB GPU.

\subsection{\textit{In-Domain} Generalization}
\texttt{AstroSpy}'s \textit{in-domain} generalization capabilities were thoroughly evaluated using a test set comprising real and synthetic astronomical images. The evaluation aimed to determine how well \texttt{AstroSpy} distinguishes between authentic and generated images within the same domain as the training data. The results, as shown in Table \ref{tab:results}, demonstrate that \texttt{AstroSpy} significantly outperforms baseline models that use only image or spectral features. \texttt{AstroSpy} achieves an accuracy of 98.5\%, highlighting the effectiveness of concatenating both feature types.
\vspace{-0.5em}
\subsection{Generalization to \textit{Out-of-Domain} Distributions}
\begin{figure}[h]
\centering
\includegraphics[width=\linewidth]{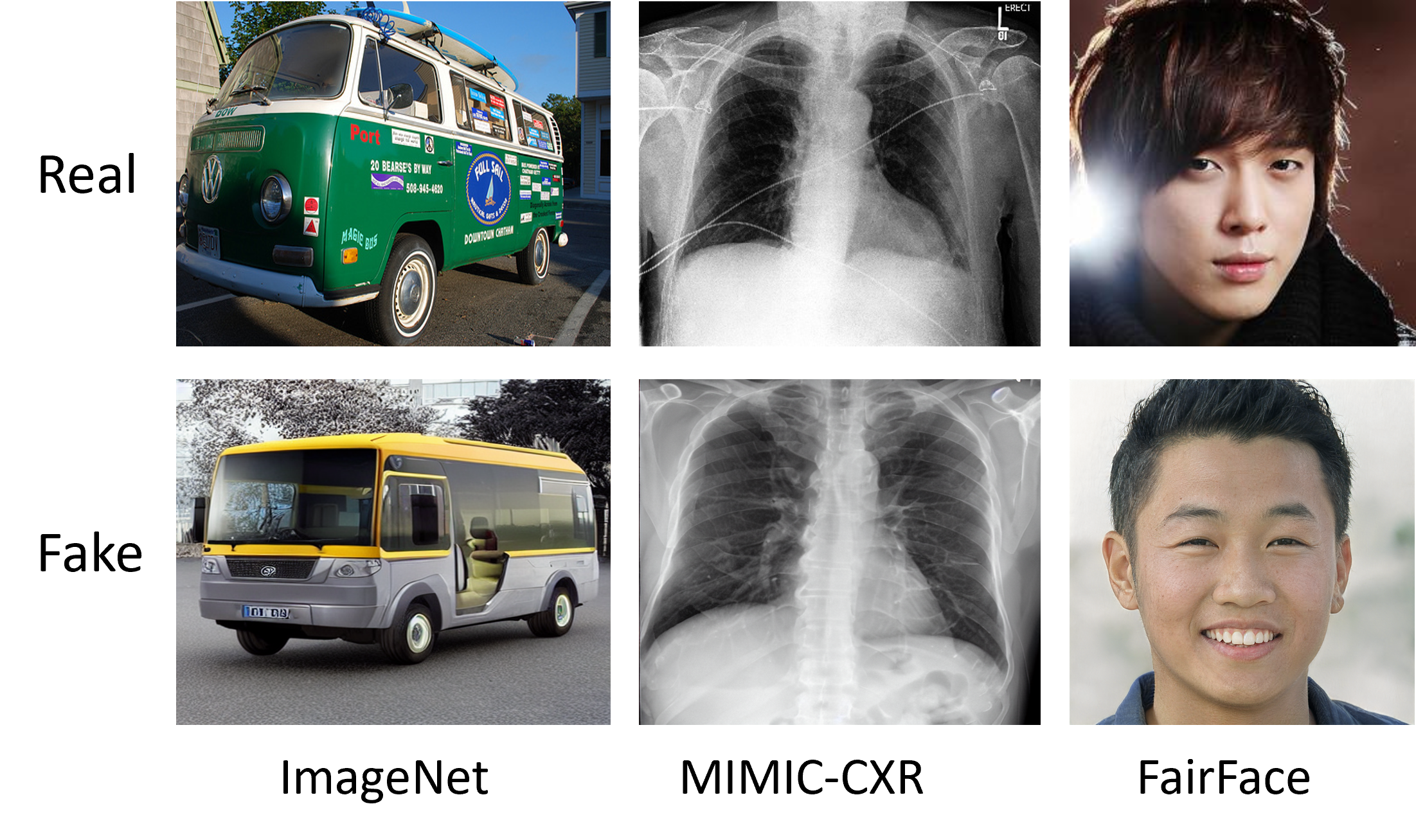}
\caption{Samples of \textit{out-of-domain} datasets used for testing \texttt{AstroSpy}'s generalization capabilities. The top row shows real images from a specific class, while the bottom row shows corresponding generated images.}
\label{fig:OOD-samples}
\end{figure}

\texttt{AstroSpy}'s ability to generalize to entirely different domains was tested using samples generated by UniDiffuser, Roentgen, and StyleGAN2. These samples include natural images, medical images, and face images from various ethnicities, respectively as shown in Figure \ref{fig:OOD-samples}. \texttt{AstroSpy} consistently outperformed baseline models as shown in Table \ref{tab:results}, demonstrating its adaptability to different types of images beyond the astronomical domain. This cross-domain robustness underscores the potential of \texttt{AstroSpy} in broader applications, including medical imaging and forensic analysis.

\begin{figure}[h]
    \centering
    \begin{minipage}[b]{0.23\textwidth}
        \centering
        \includegraphics[width=\linewidth]{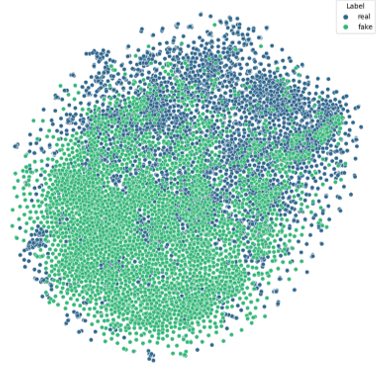}
        \caption*{(a) Baseline (bs.)} 
    \end{minipage}
    \hfill
    \begin{minipage}[b]{0.23\textwidth}
        \centering
        \includegraphics[width=\linewidth]{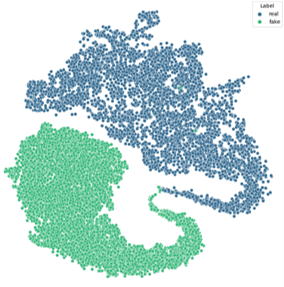}
        \caption*{(b) \texttt{AstroSpy} (Ours)}
    \end{minipage}
    \caption{t-SNE visualization of the joint embeddings from \texttt{AstroSpy}. The plot shows distinct clusters for real (blue) and fake (green) images, indicating effective feature separation and the model's ability to differentiate between real and generated astronomical images. Baseline indicates embeddings from Fourier Transformed embeddings.}
    \label{fig:tsne_plots}
\vspace{-1.0em}
\end{figure}
\vspace{-1.0em}
\subsection{Feature Shifts via \texttt{AstroSpy}}
We employ t-SNE visualizations to analyze the feature shifts captured by \texttt{AstroSpy}. The embeddings show distinct clusters for real and fake images, validating the effectiveness of our joint feature representation. Figure \ref{fig:tsne_plots} illustrates the t-SNE visualization of the joint embeddings.
The distinct separation between real and fake images in the plot underscores the discriminative power of \texttt{AstroSpy}'s joint embeddings, reinforcing the model's ability to effectively differentiate between real and generated astronomical images.
\subsection{Qualitative Analysis}
\begin{figure}[t]
\centering
\includegraphics[width=\linewidth]{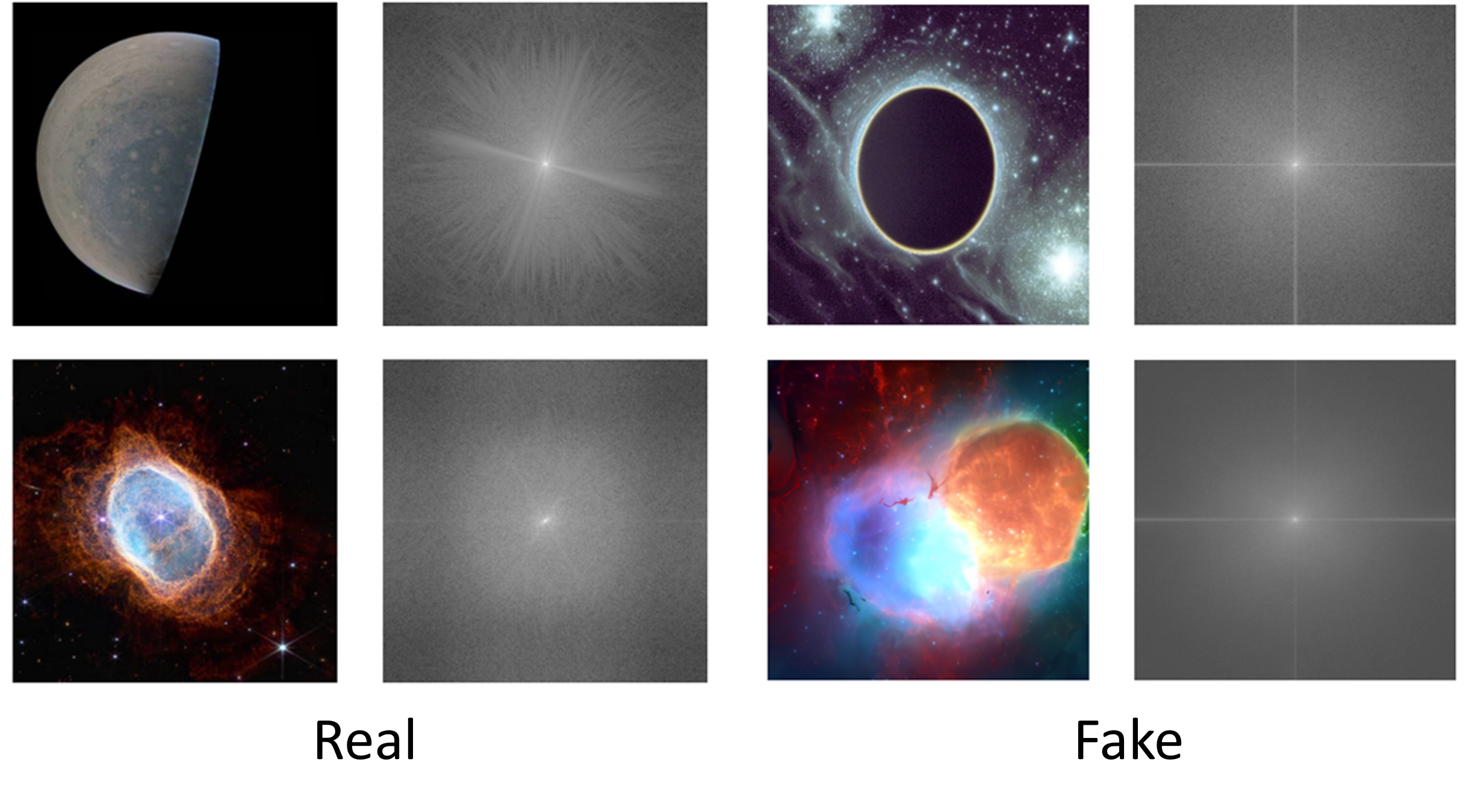}
\caption{Samples of real (left column) and synthetic (right column) astronomical images with their corresponding spectra.The spectra reveal distinct patterns that help in differentiating real images from fakes.}
\label{fig:data-samples}
\end{figure}

To understand the effectiveness of \texttt{AstroSpy}, we provide a qualitative analysis of real and synthetic images along with their spectra, as shown in Figure \ref{fig:data-samples}. Real images exhibit natural and continuous spectral patterns, indicative of genuine celestial structures. In contrast, synthetic images display spectral artifacts and frequency distributions due to generation processes.

\begin{itemize}
\item \textbf{Real Images and Spectra}: The real images show clear and detailed astronomical objects with well-defined frequency components and smooth gradients in their spectra, reflecting their authenticity.
\vspace{-1.5em}
\item \textbf{Synthetic Images and Spectra}: The synthetic images mimic real astronomical appearances but lack intricate details and natural variations. Their spectra reveal artifacts and patterns, highlighting the limitations of generation models.
\end{itemize}


\subsection{Impact of Data Augmentation}
Table \ref{tab:augmentation} summarizes the impact of different augmentation techniques on the model's performance. The results demonstrate that individual augmentations such as horizontal flip, color jitter, and random rotation significantly improve model performance compared to no augmentation. However, the combined application of all augmentation techniques yields the highest accuracy and F1-score, achieving 98\% and 97\%, respectively. 

{\renewcommand{\arraystretch}{1.0}
\begin{table}[t]
\caption{Impact of different augmentation approaches when training the \texttt{AstroSpy} detector.}
\vspace{-0.75em}
\centering
\resizebox{0.45\textwidth}{!}{%
\begin{tabular}{l|ll}
\hline
\rowcolor[HTML]{EFEFEF} 
\textbf{Augmentation Type} & \textbf{Accuracy} & \textbf{F1-Score} \\
\hline
No Augmentation & 0.88 & 0.87 \\
Horizontal Flip & 0.93 & 0.92 \\
Vertical Flip & 0.89 & 0.88 \\
Random Rotation & 0.91 & 0.90 \\
Color Jitter & 0.94 & 0.93 \\
Gaussian Blur & 0.89 & 0.88 \\
\rowcolor[HTML]{E4EDFA} 
\textbf{Combined Augs.} & \textbf{0.98}$~\uparrow$  & \textbf{0.97}$~\uparrow$  \\

\hline
\end{tabular}}
\label{tab:augmentation}
\end{table}}
\vspace{-1.2em}
\section{Conclusion}

In this study, we introduced \texttt{AstroSpy}, a hybrid model that combines spatial and spectral features to distinguish real astronomical images from fakes, demonstrating superior performance. Using a unique dataset of real and synthetic images, \texttt{AstroSpy} effectively maintains the authenticity of astronomical data. Future work will explore \texttt{AstroSpy}'s robustness against sophisticated fake image generation processes, including commercial generators in out-of-domain analysis. We will also examine the potential misuse of \texttt{AstroSpy} by malicious actors to create more convincing fakes, providing insights to strengthen detection methods and develop countermeasures. These efforts aim to enhance \texttt{AstroSpy}'s effectiveness and ensure scientific integrity and public trust.




\printbibliography
\addcontentsline{toc}{section}{References}

\end{document}